# Unitary Learning for Deep Diffractive Neural Network

**Yong-Liang Xiao**

School of Physics and Optoelectronic Engineering, Xiangtan University, Xiangtan 411105, China

ylxiao@xtu.edu.cn

**Abstract:** Realization of deep learning with coherent diffraction has achieved remarkable development nowadays, which benefits on the fact that matrix multiplication can be optically executed in parallel as well as with little power consumption. Coherent optical field propagated in the form of complex-value entity can be manipulated into a task-oriented output with statistical inference. In this paper, we present a unitary learning protocol on deep diffractive neural network, meeting the physical unitary prior in coherent diffraction. Unitary learning is a backpropagation serving to unitary weights update through the gradient translation between Euclidean and Riemannian space. The temporal-space evolution characteristic in unitary learning is formulated and elucidated. Particularly a compatible condition on how to select the nonlinear activations in complex space is unveiled, encapsulating the fundamental sigmoid, tanh and quasi-ReLu in complex space. As a preliminary application, deep diffractive neural network with unitary learning is tentatively implemented on the 2D classification and verification tasks.

## 1. Introduction

Deep learning has been received as an amazing paradigm for image recognition, language translation, computational imaging, and so on [1], it is being implemented in both academia and industry with an explosive growth. The applications strongly require ultra-fast speed and little power consumption hardware that mimicking the biological neurons and synapses [2-4], even real-time processing in some occasions. Optical neural network has been investigated for several decades within Fourier optics, forward physical architecture of multilayer coherent neural network has anticipated a promising prospect [5-6] and evolved into on-the-fly Nanophotonics [7]. Recently, deep diffraction neural network(**[DN]$^2$**) [8] has been proposed to validate classification tasks with Terahertz spectrum illumination following Huygens-Fresnel principle [9], it could approach to high dimensional information capacity with optical processing, reaching millions of neurons and hundreds of billions of connections, and implementing in Fourier space for image salient dection [10].

Diffraction is a ubiquitous physical phenomenon in coherent optical propagation. An arbitrary point located at the plane perpendicular to the propagation direction could be interpreted as a coherent superposition of complex amplitude that comes from the holistic points at the referenced plane with a certain diffractive distance. The superposition property in diffraction fulfills a fundamental requirement for deep complex neural network [11-12]. Thus, coherent diffraction, as a matter of optical mechanism, could provide an alternative manner for fully connecting complex-value neurons [13-15], tailored to be a **[DN]$^2$**. All-optical nonlinear property of photoelectric material is also being investigated as a nonlinear photoelectric synapse [16-18].

**[DN]$^2$** has the superiority of exploring the whole optical field in the propagation inducing amplitude and phase information, rather than separated real and imaginary parts. Commonly, the real or imaginary part of coherent optical field does not have any primarily substantial meaning [19], splitting the real and imaginary components into two separate real-valued channels could discard some of the complex algebraic structure of the data, does not maintain the phase information as an entity. The plausible complex-value training with the separate format in real part and imaginary part, in which the neurons in the real-part network have independent connections to those in the imaginary-part network, could present vague meaningful generalization characteristics in physical implementation [20]. Thus, a single channel complex-value training is a judicious and desirable option.

For complex-value training, a holomorphic constraint on activations in complex space applied in general complex-value backpropagation has brought about a compact formulation [21], whereas few of the minority of synapses in real space extended into complex space meets the holomorphic condition, for example, the fundamental sigmoid and tanh. Thus, it is pressing as well as urgent for us to reconcile the dilemma between mathematical definition and physical implementation for backpropagation in training. Additionally, coherent diffraction utilized in **[DN]²** has an intrinsic property that the optical field diffraction in free space is a physically unitary transformation.

Here, we formulate a *unitary learning* protocol for **[DN]²** with the help of compatible learning. Compatible learning is a general learning measure overcomes the prolonged ambiguity on the differentiability in the complex space for nonlinear synapse, enabling the fundamental sigmoid, tanh, and quasi-ReLu in complex space could be successfully implemented in backpropagation. The intrinsic unitary property in Fresnel diffraction propagation is further enforced on compatible learning. The Euclidean gradient in compatible learning is translated into Riemannian gradient, which is operated on the Lie group of unitary matrices. Additionally, the two compact formulations for complex-value backpropagation with holomorphic and compatible conditions are discussed. Considering practical implementations in **[DN]²**, a series of quasi-phase ReLu synapses are profoundly investigated for the effective training. As examples of application, we apply unitary learning in **[DN]²** to deploy 2D classification and verification tasks in unitary space.

## 2. Principle of Deep Diffraction Neural Network

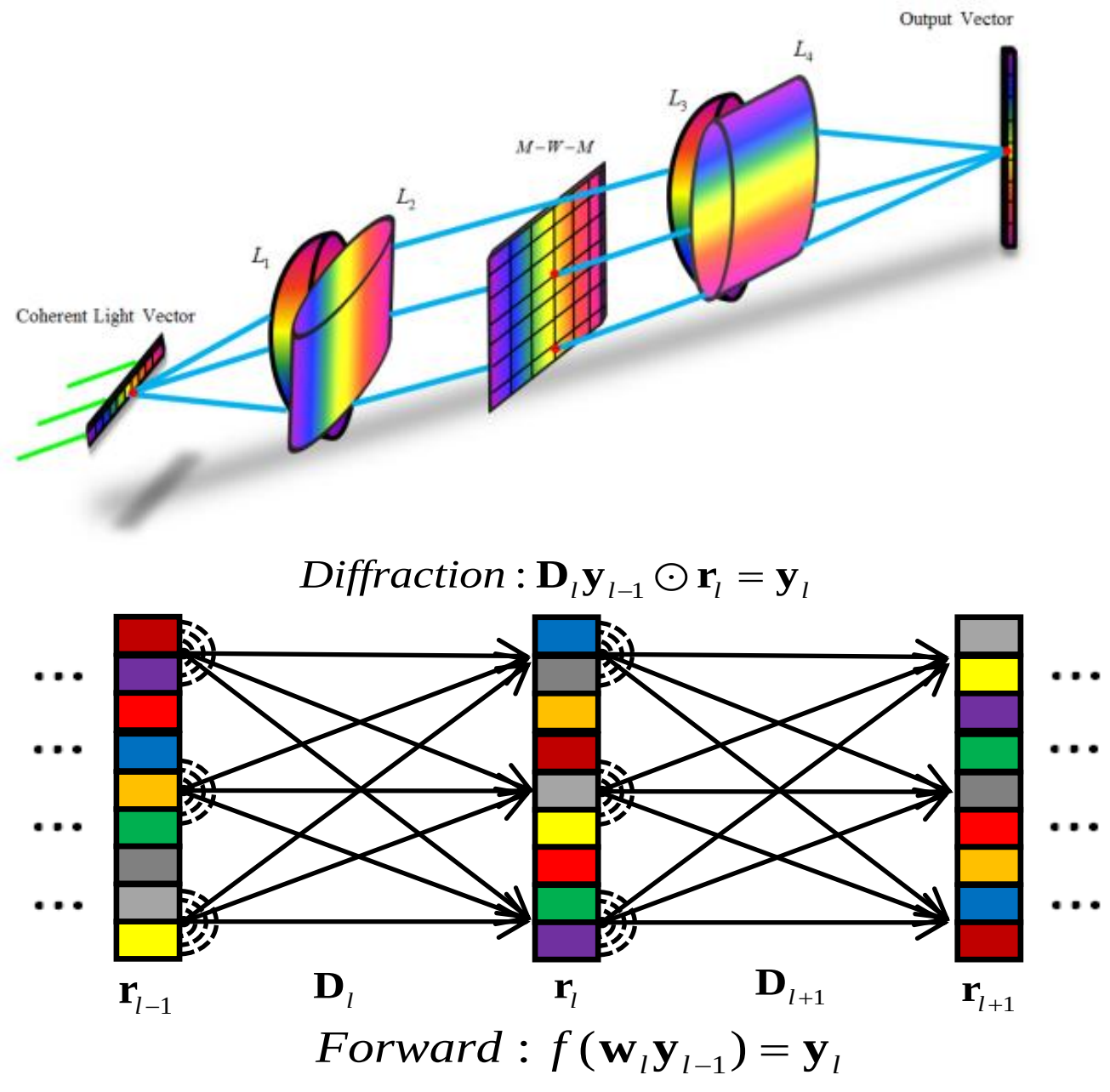

***Fig.1*** *Optical schematic for deep diffractive neural network and its analogous portrait*

**[DN]²** is build based on an analogous portrait of optical coherent diffraction. An optical node is shown in Figure 1, a synapse collects the complex-value neurons, accomplishing by diffractive coherent superposition, and is subsequently nonlinear activated with photoelectric material. Here, a coherent modulation mechanism is introduced into nodes exploiting the diffraction propagation matrix. For a multilayer mechanism, coherent diffraction and digital neural network in the $nth$ layers can be modulated to be an equivalence. $\mathbf{D}_n$ is used to represent discrete coherent diffraction matrix [22], involving Fourier、Fresnel、Fractional Fourier operator with respect to a certain diffractive distance.

*Deep Diffraction Forward mode*

$$\mathbf{D}_n(\mathbf{D}_{n-1}(\mathbf{D}_{n-2}\cdots(\mathbf{D}_2(\mathbf{D}_1\mathbf{y}_0\odot\mathbf{r}_1)\odot\mathbf{r}_2)\cdots\odot\mathbf{r}_{n-2})\odot\mathbf{r}_{n-1})\odot\mathbf{r}_n$$

*Digital Neural Network Forward mode*

$$\mathbf{W}_n f_{n-1}(\mathbf{W}_{n-1} f_{n-2}(\mathbf{W}_{n-2}\cdots f_2(\mathbf{W}_2 f_1(\mathbf{W}_1\mathbf{y}_0))\cdots))$$

The modulation mechanism in the $nth$ layer is built as

$$\begin{cases} f_1(\mathbf{W}_1\mathbf{y}_0)=\mathbf{y}_1 \\ f_2(\mathbf{W}_2\mathbf{y}_1)=\mathbf{y}_2 \\ \quad\vdots \qquad \vdots \\ f_n(\mathbf{W}_n\mathbf{y}_{n-1})=\mathbf{y}_n \end{cases} \qquad \begin{cases} \mathbf{D}_1\mathbf{y}_0\odot\mathbf{r}_1=\mathbf{y}_1 \\ \mathbf{D}_2\mathbf{y}_1\odot\mathbf{r}_2=\mathbf{y}_2 \\ \quad\vdots \qquad \vdots \\ \mathbf{D}_n\mathbf{y}_{n-1}\odot\mathbf{r}_n=\mathbf{y}_n \end{cases}$$

$$f_n(\mathbf{W}_n\mathbf{y}_{n-1})=\mathbf{D}_n\mathbf{y}_{n-1}\odot\mathbf{r}_n \qquad (1)$$

$\odot$ denotes Hadamard product. Diffraction distances are sampled as $z_n=d^2/\lambda N$ with the size in an area $d\times d$ with $N\times N$ pixels in each layer, $\lambda$ is the illuminating coherent wavelength. Coherent propagation is modulated to be an output optical signal $\mathbf{y}_n$ in $nth$ layer through nonlinear synapse $f_n(\mathbf{x})$ , of which the input signal $\mathbf{y}_{n-1}$ is multiplied by the optical intelligence matrix $\mathbf{W}_n$ learned with backpropagation. $\mathbf{r}_n$ is a forward modulation. Taking the intrinsic unitary on coherent diffraction into account, unitary constraint should be enforced on desktop backpropagation in complex space, and it would be translated from euclidean gradient in compatible learning to Riemannian gradient in unitary learning, operated on the Lie group of unitary matrices [23-24].

# 3. Compatible learning

## 3.1 Mathematical Description

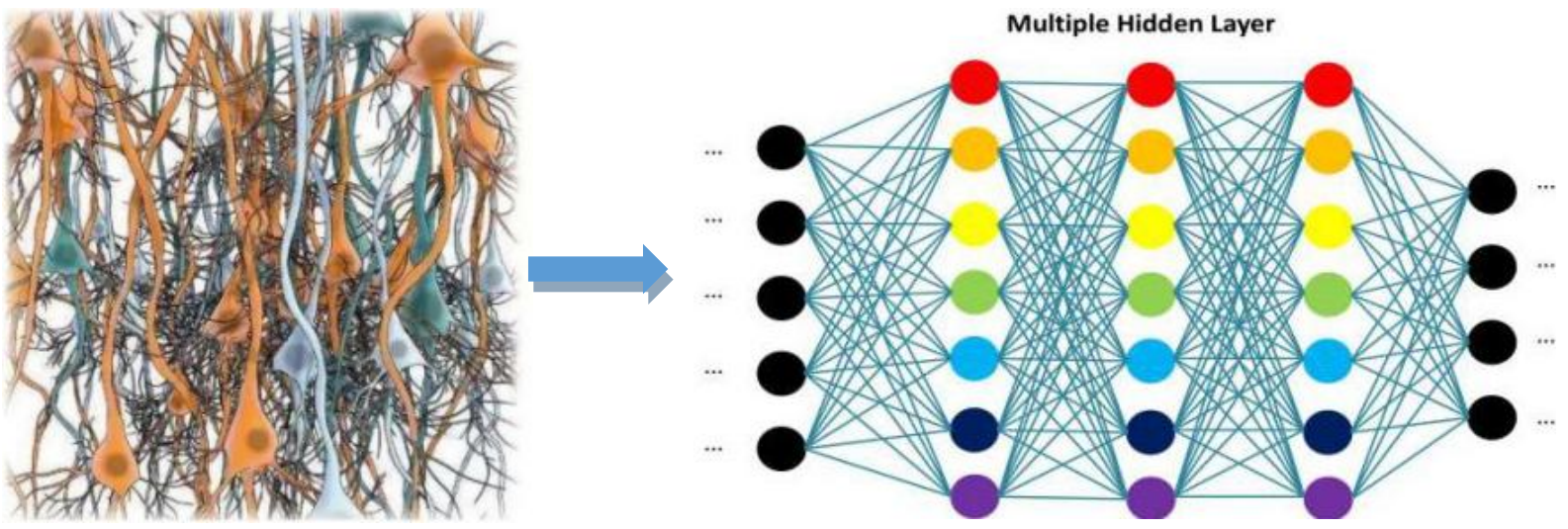

***Fig.2*** Schematic diagram of deep neural network in complex space mimicked from biological neuron

*Forward propagation* Neural network originates from neuroscience [25-26], current salient feature of deep learning is that limited success on understanding information processing through hierarchical multilayer networks under the viewpoint of complex space. A deep neural network in complex space involves many different linear combiners, each of which has a nonlinear activation at the outputs. Considering the variable as a single entity rather than bi-variate real number [19], forward propagation for a full connection in complex space is expressed as

$$\mathbf{x}_\ell = f(\mathbf{z}_\ell), \mathbf{z}_\ell = \mathbf{W}_\ell\mathbf{x}_{\ell-1}+\mathbf{b}_\ell \qquad (2)$$

*Backward Propagation* Square-error is regarded as a loss function. We implement a conjugate gradient operator for finding optimized path when process a function of the independent $\mathbf{z}$ and $\mathbf{z}^*$. The conjugate gradient operator can solve the minimization problem directly and mathematically in secure with respect to complex matrices. We can compute the complex variables derivatives of loss function with respect to the weight using complex-chain rules. Defining *kth* denotes the current layer in the multilayer deep neural network in complex space with a full connection, and holistic layer is designed to be $\ell$. The backpropagation and update in complex space for deep complex neural network, for the first time, is demonstrated as follow

$$
\begin{aligned}
\nabla_{\mathbf{w}_k^*}\boldsymbol{\varepsilon}\left[\boldsymbol{\varepsilon}^*\right] &= \boldsymbol{\varepsilon}\nabla_{\mathbf{w}_k^*}\left[\mathbf{t}_{\ell+1}^* - f(\mathbf{W}_{\ell+1}^* f(\mathbf{W}_{\ell}^* \cdots f\left[\mathbf{W}_{k+1}^* f(\mathbf{W}_k^*\mathbf{x}_{k-1}^* + \boldsymbol{\theta}_k^*) + \boldsymbol{\theta}_{k+1}^*\right] \cdots + \boldsymbol{\theta}_{\ell}^*) + \boldsymbol{\theta}_{\ell+1}^*)\right] \\
&= -\boldsymbol{\varepsilon}\nabla_{\mathbf{w}_k^*} f\left(\mathbf{net}_{l+1}^*\right)\frac{\partial\left[\mathbf{net}_{\ell+1}^*\right]}{\partial\mathbf{x}_{\ell}^*}\frac{\partial\left[\mathbf{x}_{\ell}^*\right]}{\partial\mathbf{net}_{\ell}^*}\frac{\partial\left[\mathbf{net}_{\ell}^*\right]}{\partial\mathbf{x}_{\ell-1}^*}\frac{\partial\left[\mathbf{x}_{\ell-1}^*\right]}{\partial\mathbf{net}_{\ell-1}^*}\cdots\frac{\partial\left[\mathbf{net}_{k+2}^*\right]}{\partial\mathbf{x}_{k+1}^*}\frac{\partial\left[\mathbf{x}_{k+1}^*\right]}{\partial\mathbf{net}_{k+1}^*}\frac{\partial\left[\mathbf{net}_{k+1}^*\right]}{\partial\mathbf{x}_{k}^*}\frac{\partial\left[\mathbf{x}_{k}^*\right]}{\partial\mathbf{net}_{k}^*}\frac{\partial\left[\mathbf{net}_{k}^*\right]}{\partial\mathbf{W}_{k}^*}
\end{aligned}
$$

$$
\begin{aligned}
\Delta\mathbf{W}_k &= -\mu\left\{\mathbf{W}_{k+1}^H\mathbf{W}_{k+2}^H\cdots\mathbf{W}_{\ell}^H\mathbf{W}_{\ell+1}^H\right\}\left\{\left[(-\varepsilon)\odot f'\left(\mathbf{net}_{\ell+1}^*\right)\right]\odot f'\left(\mathbf{net}_{\ell}^*\right)\odot f'\left(\mathbf{net}_{\ell-1}^*\right)\cdots\odot f'\left(\mathbf{net}_{k+1}^*\right)\odot f'\left(\mathbf{net}_{k}^*\right)\right\}\mathbf{x}_{k-1}^H \\
&= \mu\boldsymbol{\varepsilon}\left[\prod_{t=k+1}^{\ell+1}\mathbf{W}_t^H\right]\left[\Theta_{t=k}^{\ell+1} f'\left(\mathbf{net}_t^*\right)\right]\mathbf{x}_{k-1}^H
\end{aligned}
$$

$$
\boldsymbol{\delta}_{\ell+1} = (-\boldsymbol{\varepsilon})\odot f'\left(\mathbf{net}_{\ell+1}^*\right), \boldsymbol{\delta}_{\ell} = \mathbf{W}_{\ell+1}^H\boldsymbol{\delta}_{\ell+1}\odot f'\left(\mathbf{net}_{\ell}^*\right), \boldsymbol{\delta}_{\ell-1} = \mathbf{W}_{\ell}^H\boldsymbol{\delta}_{\ell}\odot f'\left(\mathbf{net}_{\ell-1}^*\right)\cdots
$$

$$
\boldsymbol{\delta}_{k+1} = \mathbf{W}_{k+1}^H\boldsymbol{\delta}_{k+2}\odot f'\left(\mathbf{net}_{k+1}^*\right), \boldsymbol{\delta}_{k} = \mathbf{W}_{k}^H\boldsymbol{\delta}_{k+1}\odot f'\left(\mathbf{net}_{k}^*\right)
$$

$$
\Delta\mathbf{W}_k = -\mu\boldsymbol{\delta}_k\mathbf{x}_{k-1}^H
$$

**Real Space**

$$
\mathbf{x}_{\ell} = f(\mathbf{u}_{\ell}), \mathbf{u}_{\ell} = \mathbf{W}_{\ell}\mathbf{x}_{\ell-1} + \mathbf{b}_{\ell}
$$

$$
\begin{cases}
\boldsymbol{\delta}_L = (\mathbf{u}_L - \mathbf{t}_L)\odot f'(\mathbf{u}_L) \\
\boldsymbol{\delta}_{\ell} = \left[\mathbf{W}_{\ell+1}\right]^T\boldsymbol{\delta}_{\ell+1}\odot f'(\mathbf{u}_{\ell}) \\
\Delta\mathbf{W}_{\ell} = -\eta\boldsymbol{\delta}_{\ell}\left[\mathbf{x}_{\ell-1}\right]^T \\
\Delta\mathbf{b}_{\ell} = -\eta\boldsymbol{\delta}_{\ell}
\end{cases}
$$

$$
\mathbf{x}_{\ell}, \mathbf{x}_{\ell-1}, \mathbf{u}_{\ell}, \mathbf{u}_L, \mathbf{t}_L, \boldsymbol{\delta}_L, \boldsymbol{\delta}_{\ell}, \Delta\mathbf{W}_{\ell}, \Delta\mathbf{b}_{\ell} \in \Re
$$

**Complex Space**

$$
\mathbf{x}_{\ell} = f(\mathbf{z}_{\ell}), \mathbf{z}_{\ell} = \mathbf{W}_{\ell}\mathbf{x}_{\ell-1} + \mathbf{b}_{\ell}
$$

$$
\begin{cases}
\boldsymbol{\delta}_L = (\mathbf{z}_L - \mathbf{t}_L)\odot f'(\mathbf{z}_L^*) \\
\boldsymbol{\delta}_{\ell} = \left[\mathbf{W}_{\ell+1}^*\right]^T\boldsymbol{\delta}_{\ell+1}\odot f'(\mathbf{z}_{\ell}^*) \\
\Delta\mathbf{W}_{\ell} = -\eta\boldsymbol{\delta}_{\ell}\left[\mathbf{x}_{\ell-1}^*\right]^T \\
\Delta\mathbf{b}_{\ell} = -\eta\boldsymbol{\delta}_{\ell}
\end{cases}
$$

$$
\mathbf{x}_{\ell}, \mathbf{x}_{\ell-1}, z_{\ell}, z_L, \mathbf{t}_L, \boldsymbol{\delta}_L, \boldsymbol{\delta}_{\ell}, \Delta\mathbf{W}_{\ell}, \Delta\mathbf{b}_{\ell}, \in \mathbb{Z}
$$

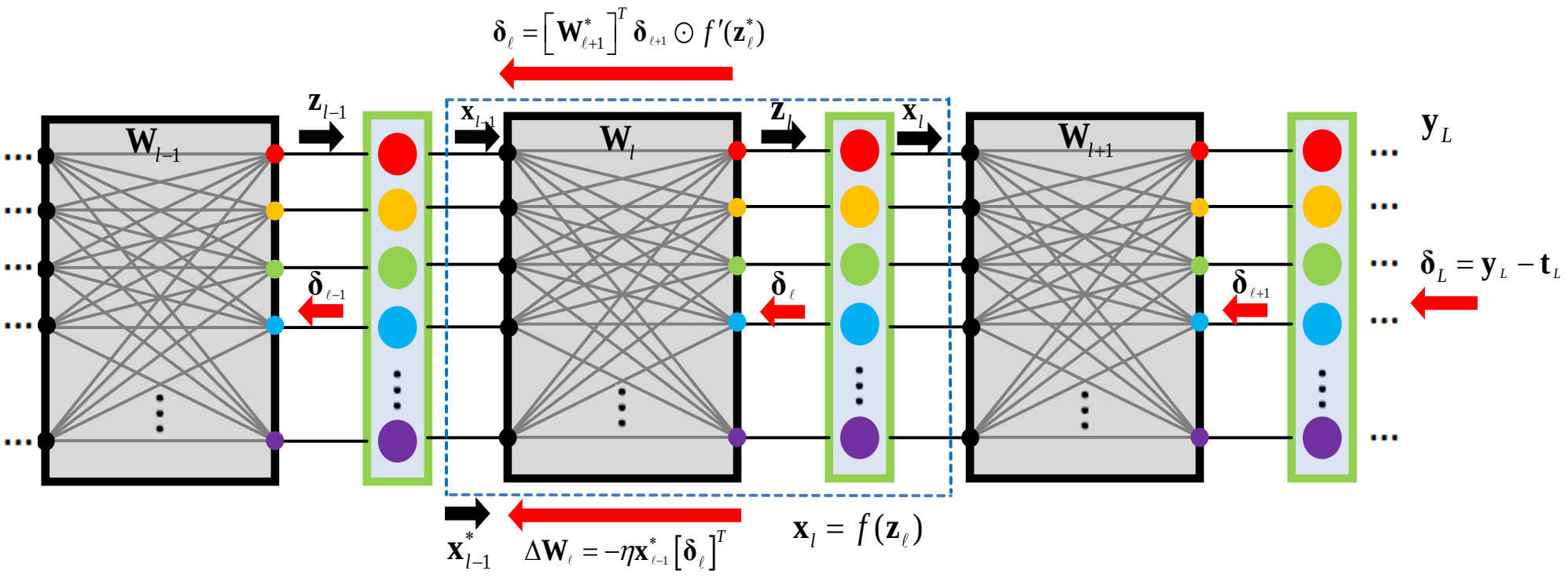

***Fig.3*** Schematic diagram of backpropagation in complex space

The learning exploited separate-type in real and imaginary parts with real-value backpropagation [26] is abandoned here, since the phase information is always propagated as an entity in optical processing implementation. Seen from the backpropagation formulation comparisons between real space and complex space, matrix representation in complex space covers its real counterpart. The update rules involving nonlinear activations for backpropagation also can degrade to be its real counterpart without any ambiguity. We term it the *compatibility*. The compatible condition, along with compatible learning during partial derivation with conjugate gradient operator, is $f^*(\mathbf{z}_{\ell}) = f(\mathbf{z}_{\ell}^*)$. Under the compatible condition, the prevalent sigmoid and tanh in real space, could be extended to its complex space by *artificially* substituting real-variable with conjugated complex-variable, encapsulating a few elementary activation functions in complex space. It is regretfully unveiled for the first time, to the best of our knowledge, furnishing a valid compact formulation without any computational ambiguity. The essence of compatible learning could be realized with the original executable code in real space just considering the phase-conjugation significance [27], carrying out complex-value initialization. The schematic diagram of compatible learning is exhibited in Figure. 3. The numerical convergence performance will be investigated with a simple training validation in phase logic.

Compatible learning is not deduced from the holomorphic condition, whilst holomorphic condition [28] could be employed to deduce a resembled matrix representation, the error update based on holomorphic condition enforced on activations is

$$\boldsymbol{\delta}_\ell = \left[\mathbf{W}^*_{\ell+1}\right]^T \boldsymbol{\delta}_{\ell+1} \odot \left[f_h^{'}(\mathbf{z}_\ell)\right]^* \qquad (3)$$

$f_h$ is a holomorphic activation function. What a pity, sigmoid and tanh, directly represented in complex space isn't a holomorphic function, can't be utilized as nonlinear activations under holomorphic condition [21]. However, sigmoid and tanh could be utilized directly as nonlinear synapses in complex space under the compatible condition. $f'(\mathbf{z}^*_\ell)$ is a function that real variable is directly replaced by the conjugation of complex variable after real derivation, while $\left[f_h^{'}(\mathbf{z}_\ell)\right]^*$ is a function that it is derived on the analytical meaning in complex space. Practically, a holomorphic condition presents some limitations for efficient implementation since, there is scarcely any prevalent activations meeting the holomorphic condition. For phase ReLu [12], $\left[f'(\mathbf{z}^*_\ell)\right]=\left[f_h'(\mathbf{z}_\ell)\right]^*$, the two models for node errors $\boldsymbol{\delta}_\ell$ under the compatible condition and the holomorphic condition has an equivalence. So does SIREN [29].

### 3.2 Nonlinear activations implemented in compatible learning

The specific matrix representation for complex-value backpropagation in full connections depends on the concrete nonlinear activations enforced on the generalized formulation. There has been a holomorphic condition that induces a concise matrix representation. However, the introduction of nonlinear activations in complex space has a notorious problem in the differentiability sothat, the prevalent sigmoid and tanh extended directly to complex space doesn't meet the holomorphic condition, sigmoid and tanh in complex space must be abandoned under the holomorphic condition. The compatible condition $f(\mathbf{z}^*_\ell) = f^*(\mathbf{z}_\ell)$, involving the fundamental sigmoid, tanh, and quasi-Relu in complex space with the concept of complex-variable conjugation substitution, can be available in compatible learning, it is paramount significance for weight update in backpropagation. The compatible condition reconciles ambiguity on nonlinear activations selection in complex space for deep complex neural network. For simplification, we utilize a 4×4 Exclusive-OR(XOR) logic training task using compatible learning [15]. Seen from Figure 4, ReLu in complex space, whose phase distributes within a half space, is fabulous for fast convergence. The learned weights for the two-layer neural network are demonstrated in Figure 5. The amplitude in the first layer is much larger than the second layer.

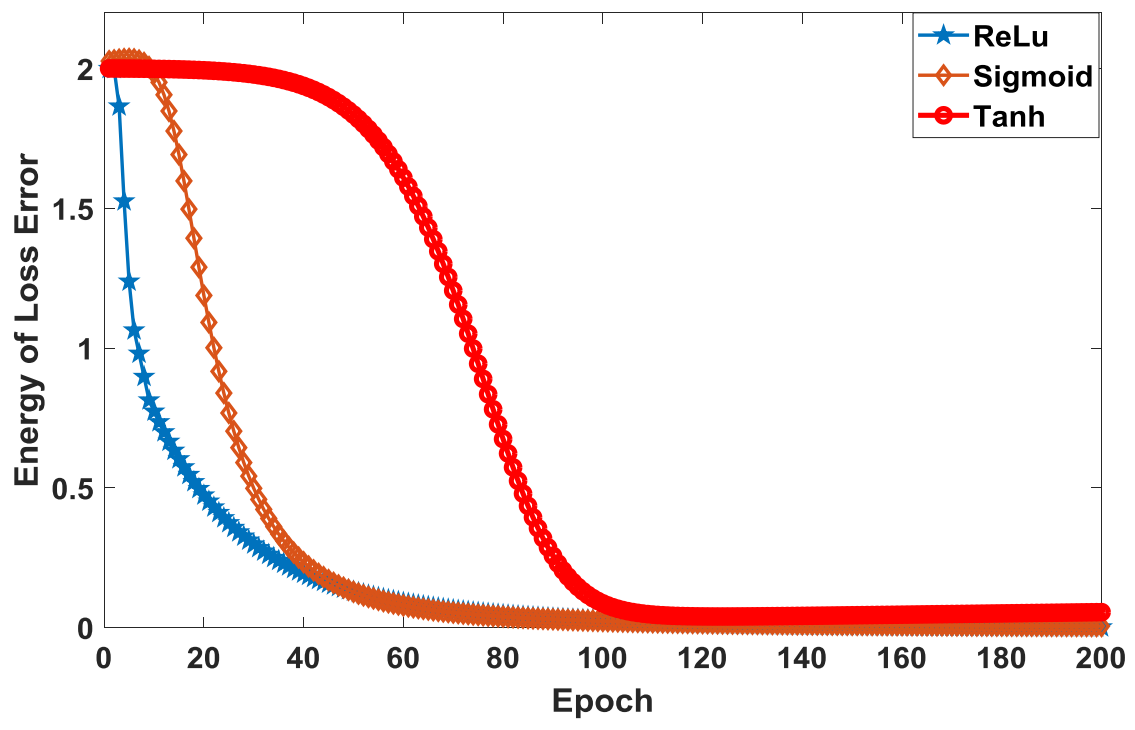

***Fig.4*** The convergent performance using nonlinear activations in complex space on 4×4 XOR logic

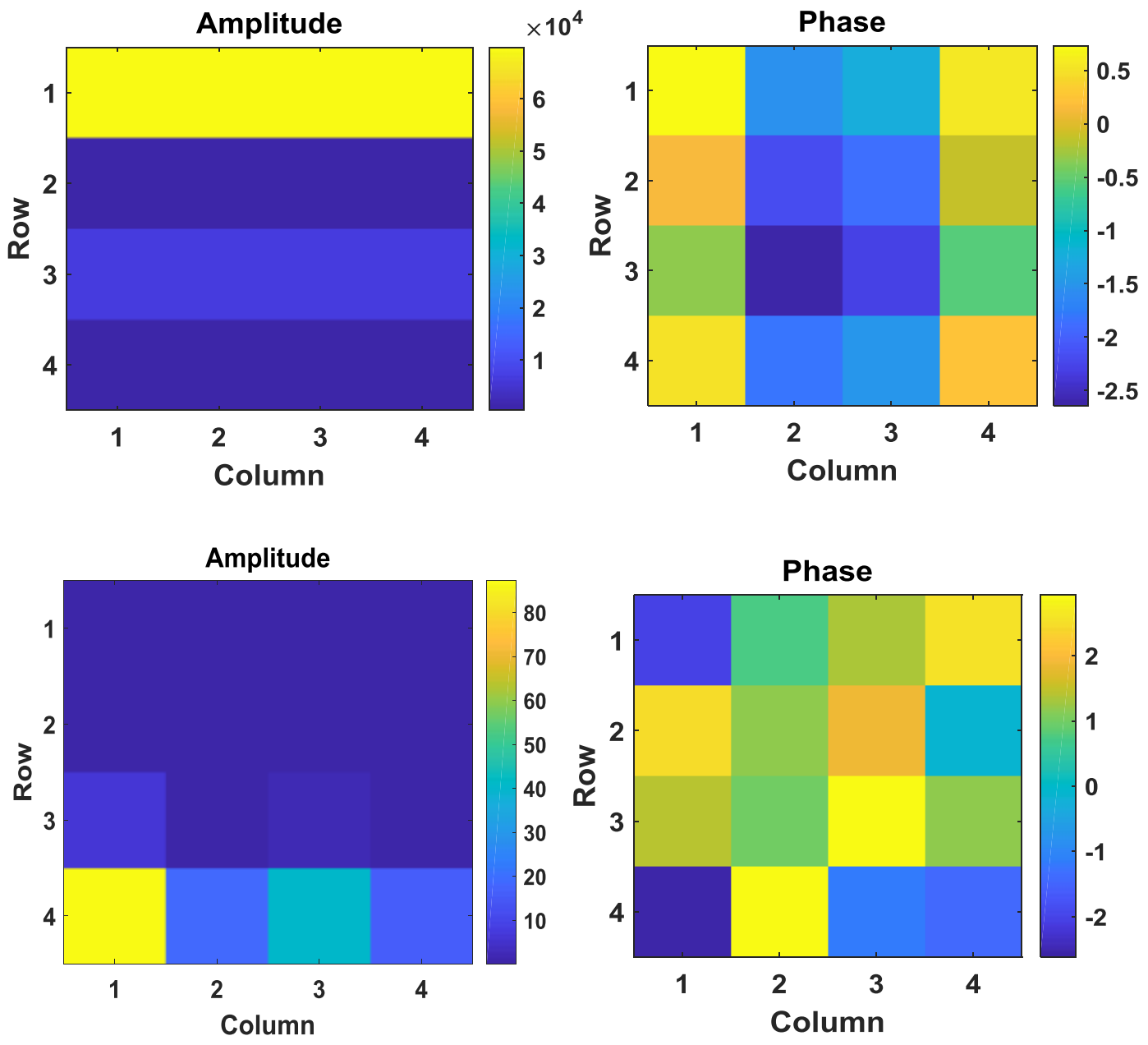

***Fig.5*** The amplitude and phase of weight within two-layer compatible learning implemented for a 4×4 XOR logic task when tanh in complex space is implemented as a synapse.

# 4. Unitary learning

## 4.1 Mathematical Description

Focusing on the problem of optimization enforced unitary constraint on the complex weight bank in deep complex neural network, it can be resolved efficiently in an elegant manner by using Riemannian geometry. Optimization for a real-valued cost function of complex-valued matrix under the unitary constraint is operated on the Lie group of unitary matrices by conjugate gradient algorithm exploiting the group properties [23-24]. Based on compatible learning, backpropagation and weight update in unitary space, for the first time, is reformulated as follows

**Unitary learning**

$$\mathbf{x}_\ell = f(\mathbf{z}_\ell), \mathbf{z}_\ell = \mathbf{U}_\ell \mathbf{x}_{\ell-1} + \mathbf{b}_\ell$$

$$\text{s.t} \quad \mathbf{U}_\ell \mathbf{U}_\ell^H = \mathbf{U}_\ell^H \mathbf{U}_\ell = \mathbf{I}$$

$$\begin{cases} \boldsymbol{\delta}_\ell = \mathbf{U}_{\ell+1}^H \boldsymbol{\delta}_{\ell+1} \odot f'(\mathbf{z}_\ell^*), \Delta\mathbf{W}_\ell = -\mu \boldsymbol{\delta}_\ell \mathbf{x}_{\ell-1}^H \\ \mathbf{G}_\ell = \mathbf{U}_\ell \Delta\mathbf{W}_\ell^H - \Delta\mathbf{W}_\ell \mathbf{U}_\ell^H \\ \Delta\hat{\mathbf{U}}_\ell = \exp\left[-\lambda \mathbf{G}_\ell\right] \\ \Delta\mathbf{b}_\ell = -\mu \boldsymbol{\delta}_\ell \end{cases}$$

$$\mathbf{x}_\ell, \mathbf{x}_{\ell-1}, \mathbf{z}_\ell, \boldsymbol{\delta}_\ell, \mathbf{U}_\ell, \Delta\mathbf{W}_\ell, \Delta\hat{\mathbf{U}}, \mathbf{G}_\ell, \Delta\mathbf{b}_\ell, \in \mathbb{Z}$$

Unitary learning is built on the compatible condition $f(\mathbf{z}_\ell^*) = f^*(\mathbf{z}_\ell)$. The compatible learning is an efficient backpropagation in complex space, providing gradient transition between euclidean space and Riemannian space for unitary learning. Gradient translation is briefly described in Fig.6. The weight update with unitary constraint is performed in the form of exponential map multiplication in Riemannian space. The meshing grid of weight update for temporal-space evolution characteristics in unitary learning is exhibited in Figure 7. The gradient translating between Riemannian and Euclidean space is taken place in temporal axis. The temporal axis with respect to optical layers denotes the iterative epoch. The meshing grid could demonstrate the procedure of unitary learning profoundly.

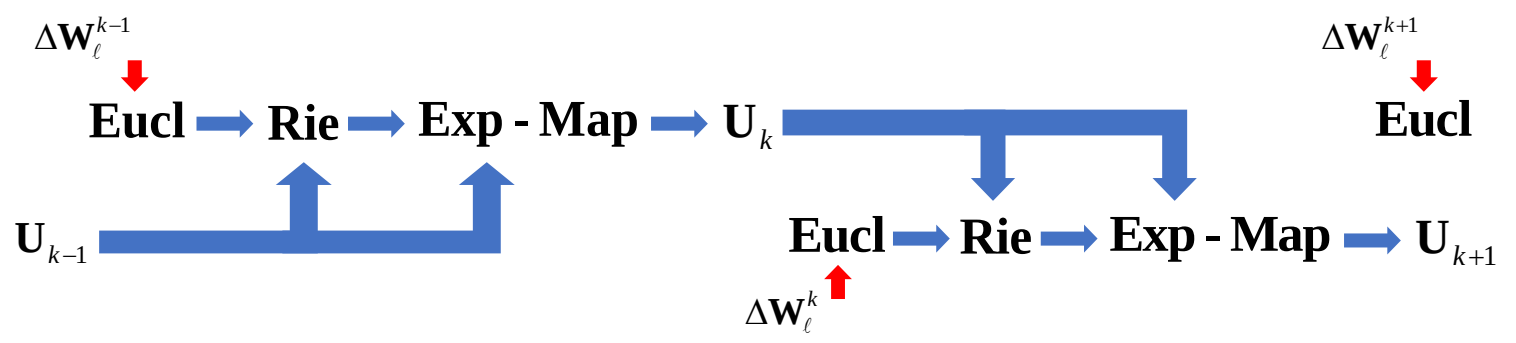

***Fig.6*** Gradient translation between Riemannian and Euclidean Space in unitary learning

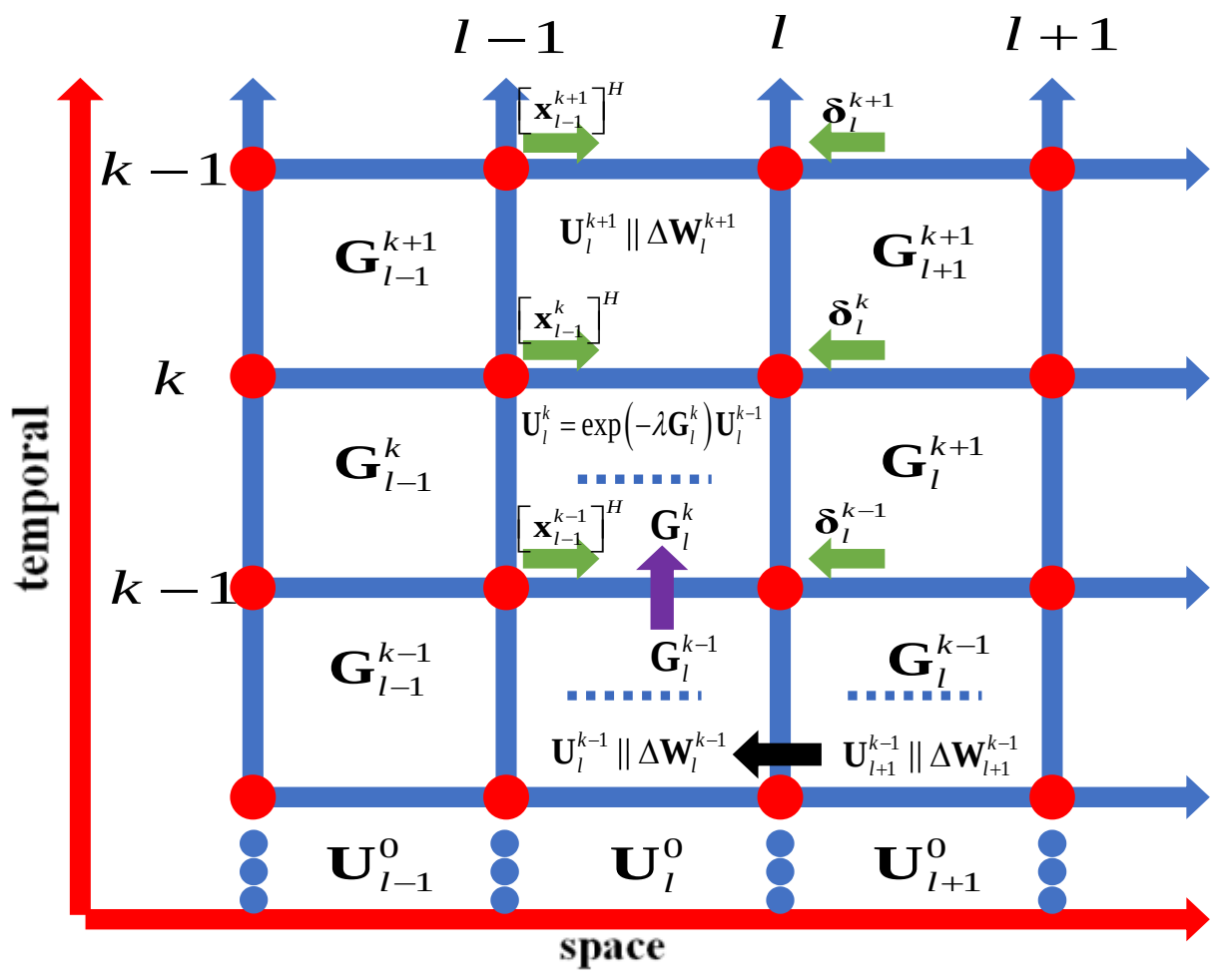

***Fig.7*** Temporal-space evolution in Unitary learning

In the grid of $\ell th$ layer, $kth$ epoch, representing as $[\ell, k]$, the unitary weight is refined by right multiplication of unitary matrix in $[\ell, k-1]$ on the exponential map $\exp[-\lambda \mathbf{G}_\ell]$ in $[\ell, k]$. The Riemannian gradient $\mathbf{G}_\ell^k$ is synthesized from the unitary weight $\mathbf{U}_\ell$ and the corresponding euclidean weight update $\Delta\mathbf{W}_\ell$ in the $[\ell, k-1]$, whose Euclidean gradient is backpropagated from the adjacent space axis $[\ell+1, k-1]$. In unitary learning, the euclidean gradient is fed by error backpropagation generated in compatible learning. The Riemannian gradient in $\ell th$ layer, $kth$ epoch for deep complex neural network is

$$\begin{cases} \Delta\mathbf{W}_\ell^k = -\boldsymbol{\varepsilon}\left[\prod_{t=\ell}^{L+1} \mathbf{U}_t^H\right]\left[\Theta_{t=\ell}^{L+1} f'\left(\mathbf{net}_t^*\right)\right]\mathbf{x}_{\ell-1}^H \\ \mathbf{G}_\ell^k = \mathbf{U}_\ell^{k-1}[\Delta\mathbf{W}_\ell^k]^H - \Delta\mathbf{W}_\ell^k\left[\mathbf{U}_\ell^{k-1}\right]^H \end{cases} \quad (4)$$

## 4. 2 Nonlinear phase synapses implemented in unitary learning

Presently, synapse ReLu in real space has been invasively utilized in deep neural network. Here, ReLu in phase space will be, for the first time, reformulated and implemented in unitary learning. Three types of quasi-phase ReLu are defined as Eq. (5). In order to test the validity of unitary learning implementing quasi-phase ReLu, we numerically perform the training of a phase logical XOR gate, which is a resemblance with a common logical XOR by unitary weights initialization. The samples shown in Table 1, are cultivated as seeds to phase logical XOR for training a $4 \rightarrow 4$ mapping with unitary weights and quasi-phase ReLu, the mapping in complex plane is demonstrated in Figure 8. The digital experiments validate unitary learning for a perfect single channel training, even both real-value and complex-value are mingled together as the training samples.

$$\mathrm{Re}\,Lu(z)=\begin{cases}|z|\mathrm{e}^{i\varphi},\text{if } \varphi\in[-\pi/2,\pi/2]\\ 0 \qquad\qquad\quad otherwise\end{cases}$$
$$L\,\mathrm{Re}\,Lu(z)=\begin{cases}|z|\mathrm{e}^{i\varphi},\text{if } \varphi\in[-\pi/2,\pi/2]\\ k|z|\mathrm{e}^{i\varphi} \qquad otherwise\end{cases} \quad (5)$$
$$E\,\mathrm{Re}\,Lu(z)=\begin{cases}|z|\mathrm{e}^{i\varphi},\text{if } \varphi\in[-\pi/2,\pi/2]\\ k(\mathrm{e}^{z}-1) \qquad otherwise\end{cases}$$

Table 1 The training samples(X | Y) for phase logical XOR gate (amplitude1 =2, amplitude2 =1)

| Samples | X1 | X2 | X3 | X4 | Y1 | Y2 | Y3 | Y4 |
|---|---|---|---|---|---|---|---|---|
| Phase1 | π/4 | 3π/4 | 5π/4 | 7π/4 | 0 | π/2 | π | 3π/2 |
| Phase2 | π/3 | 5π/6 | 8π/6 | 11π/6 | π/6 | 4π/6 | 7π/6 | 10π/6 |

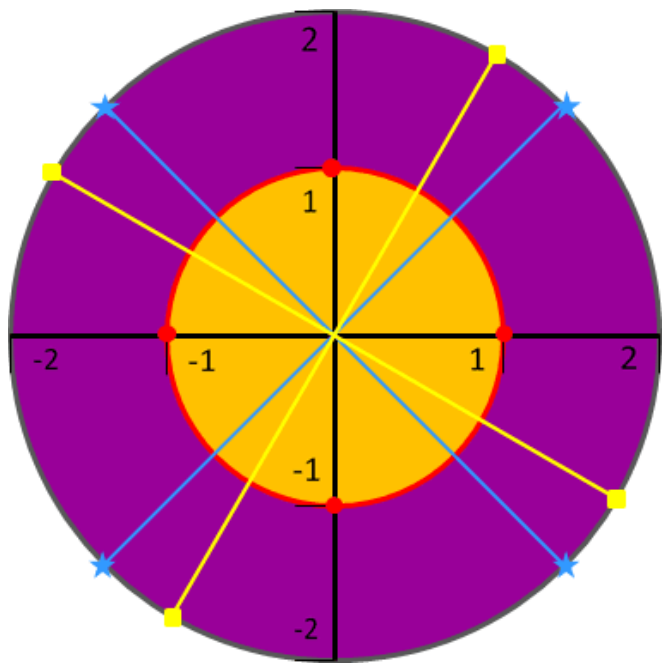

***Fig.8*** Phase logical XOR exhibition in a complex plane

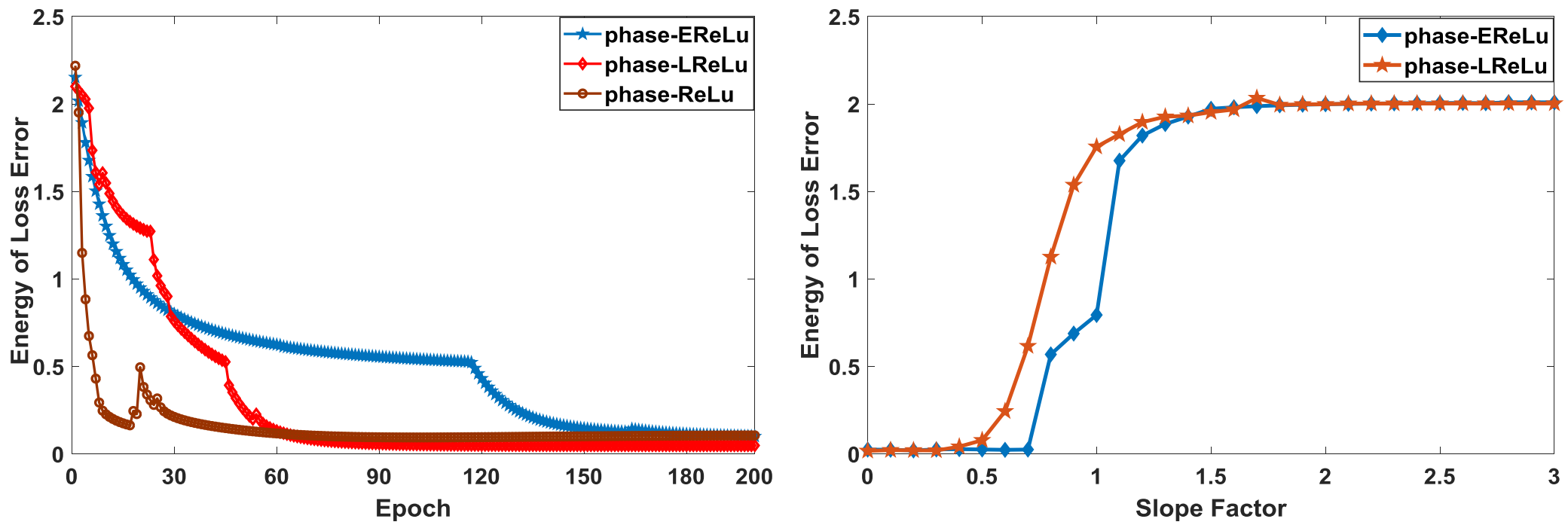

***Fig. 9*** (**a**) The convergent performance using qausi-phase-ReLu synapses on phase XOR logic

(**b**) The convergent enhancement using slope tuning factor in phase-Leaky ReLu and phase-Exponent ReLu

The corresponding learning curves are exhibited in Figure 9 when three types of quasi-phase ReLu are utilized. Seen from Figure 9(a), the convergent speed that a standard phase ReLu is applied, outperforms the others. However, the convergent performance presents a little far from perfect convergence when the slope factor is randomly chosen with the same value for phase-Leaky ReLu and phase-Exponent ReLu. In fact, the slope factor introduced to tune the standard phase ReLu could enhance the convergent performance, we further exhibit how the slope factors affect the performance when unitary learning uses phase-Leaky ReLu and phase-Exponent ReLu. Seen from Figure 9(b), the slope factor ranging from $10^{-0.5}$ to 1 can enhance the convergent performance in a specific training case with random unitary initialization. The optimal slope factor could be searched out for good convergence. Thus, phase ReLu with slope tune is very good for unitary learning.

## 5. Unitary Learning for Deep Diffractive Neural Network

Coherent diffraction AI has a strong potential for optical neural network full-connection due to its coherent superposition and convolution property. Deep neural network based on free-space diffraction has became a highly tropic for dense intelligent interconnection using coherent propagation. However, phase lost phenomena always occurs in coherent recording by a CCD sensor without any optical interference [10]. Unitary learning is considerably suitable for training diffractive neural network since complex-value represents the natural entity of optical wave under unitary diffractive transformation, even just optical intensity is directly recorded in the last layer without optical interference, allowing for statistical viewpoint rather than phase iteration [30].

### 5.1 Numerical Experiments for diffractive pattern prediction

**[DN]$^2$** could be realized following a procedure of learning complex-valued matrix herein modulation. However, coherent diffraction is a standard unitary transformation, if it is optically implemented, the intelligent matrix should be enforced a unitary constraint [31-33]. Additionally, the learned intelligent matrix is commonly very difficult to be directly mimicked with optical diffraction provided that phase-mask modulation wasn't applied. The nonlinear response in a specific layer with optical intelligence matrix and nonlinear synapses, could not be emulated just by a single optical diffraction layer. Researchers must build the relationship between optical intelligent matrix and coherent forward propagation in **[DN]$^2$**. We have given a comparison diagram in Figure 1 for describing the coherent modulation in **[DN]$^2$**, enabling forward propagation consistent with unitary learning. All the parameters are passive and fixed once the network is trained successfully by unitary learning, whereas the learned parameters are different from initialization. The random modulation for a single optical layer can be obtained with $\mathbf{r}_n = \frac{f_n(\mathbf{W}_n \mathbf{y}_{n-1})}{\mathbf{D}_n^{\varepsilon} \mathbf{y}_{n-1}}$.

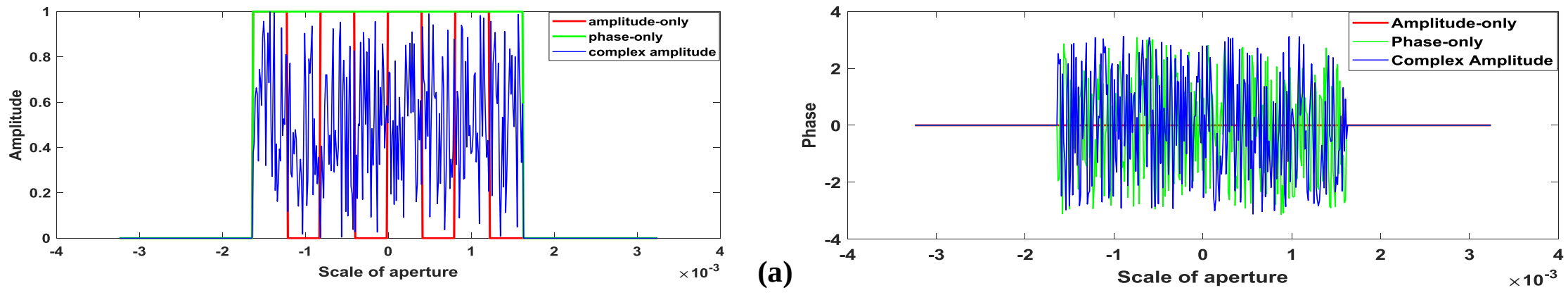

(a)

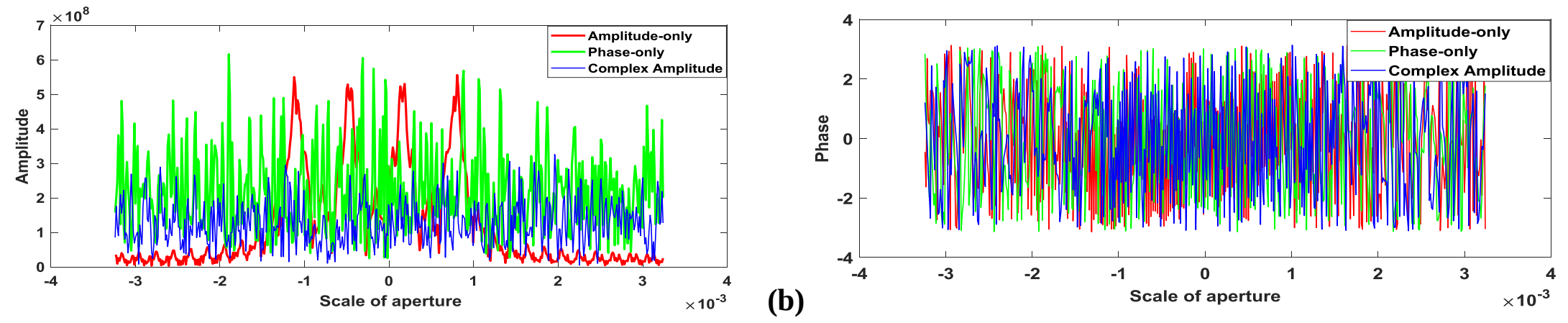

(b)

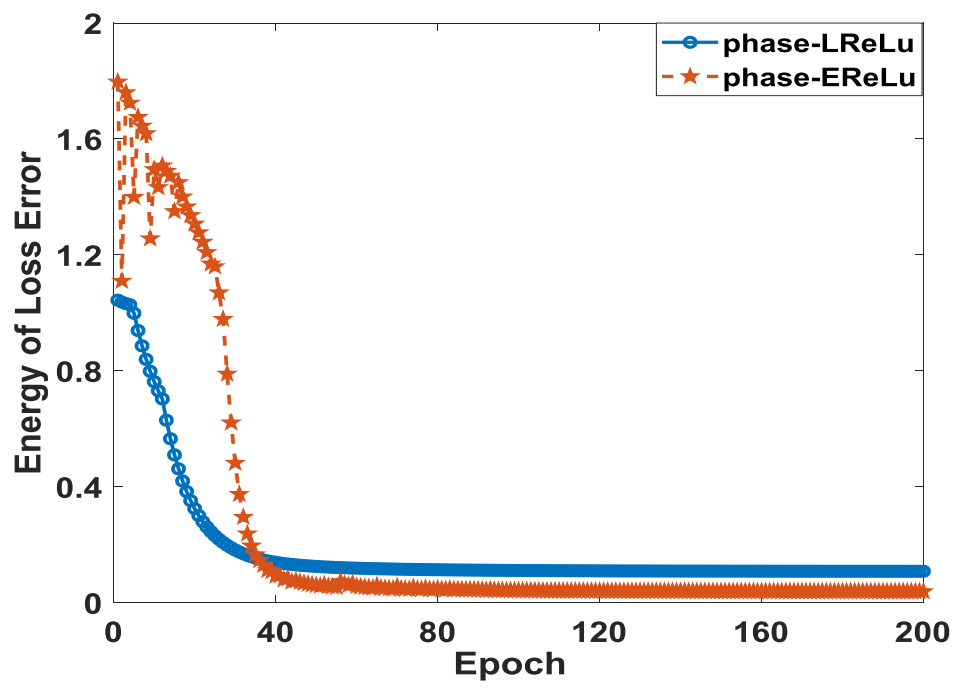

(c)

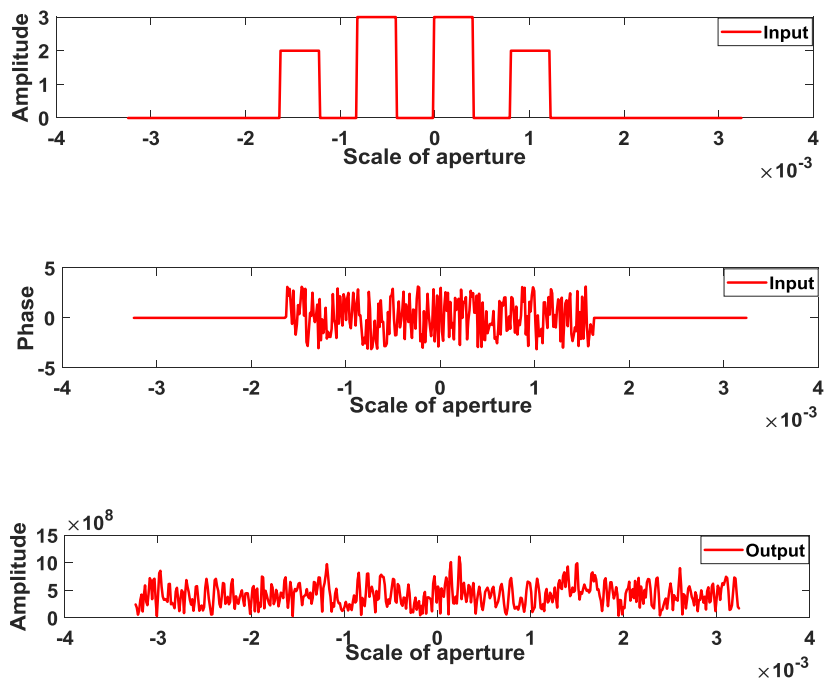

(d)

***Fig.10*** **(a)** Three types of the diffractive feeding samples with different quadrate aperture, **(b)** The corresponding diffractive samples of (a) for output layer, **(c)** The learning curves with phase-LReLu and phase-EReLu synapses, **(d)** The forward outputs for test sample using the trained deep complex neural network.

In the stage of training, optical intelligence matrix could be supervisly learned by feeding input-output optical field training sets, which are generated with Angular Spectrum Propagation (ASP) [9]. We feed training samples in two-layer **[DN]$^2$** with random modulation depicted in Figure 1, 1D signal is described here for simplicity. The number of training and validation samples are 50000, and the tested samples are 1000. In Figure 10(a)-(b), we show three types of samples involving amplitude-only, phase-only, amplitude-phase, and the corresponding diffractive coherent optical field with ASP. The scale of the samples is 5mm with 512 sampling points under 632.8nm coherent illumination, and the scale of learned unitary matrix is set as 512×512 given at 0.78m diffractive distance. The mingled training samples consist of both real-value and complex-value can also be feasible within unitary learning because of its compatibility. The learned weights could be inferred from the diffractive modulation mechanism.

For facilitating the coherent modulation, we initialize the intelligent weight with pure-imaginary unitary matrix to drive the network, making the modulations just happen at the imaginary axis in the complex plane, namely phase-shift in the modulation are $-\pi \, or \, \pi$. There is a peculiar phenomenon should be addressed during unitary learning in such a event that, the convergence never happens if the phase interval is closed at the boundary for the phase ReLu, when the pure imaginary part is provided as the driven weight. Thus, phase rotation factor $\varepsilon$ introduced into phase-step as $\left[-\pi/(2-\varepsilon), \pi/(2-\varepsilon)\right]$ could ensure an open interval characteristic in the phase boundary, or else the learning here is collapsed. Furthermore, there is a miraculous phase breakpoint unintelligibly arising in phase rotation for convergence guarantee in the event of pure-imaginary initialization, shown in Figure 11. As far as we known, it is discovered for the first time in phase ReLu. As to the phase breakpoint, the negative power for $\varepsilon$ is 16, and the coefficient is 1.1. The learning would collapse once $\varepsilon$ is lower than the phase breakpoint in a practical learning.

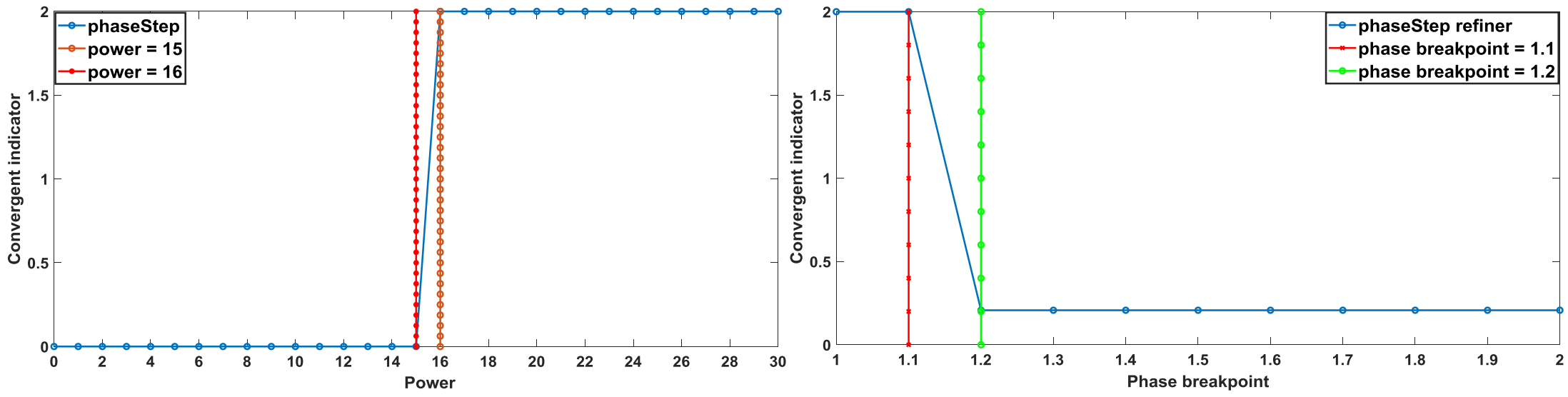

***Fig.11*** **(a)** The negative power of $\varepsilon$ for convergent indicator with imaginary initialization in Unitary learning (**b**) The coefficient of $\varepsilon$ for convergent indicator with pure-imaginary initialization in Unitary learning.

Enhanced by Leaky ReLu in phase space, the slope $k$ further introduced into the standard phase ReLu ranging from $\left[-\pi/(2-\varepsilon), \pi/(2-\varepsilon)\right]$ could accelerate and improve the convergence. The negative power of $\varepsilon$ for anti-clockwise and clockwise rotation could present a little difference. In order to investigate how the negative power of $\varepsilon$ and slope power of $k$ synergistically enhance the convergent performance, we respectively draw 3D mesh figures with clockwise and anti-clockwise rotation in Figure 12 to describe the miraculous ability of slope factor $k$, in which the coefficient of $\varepsilon$ is set as $1.1$. The adjustment of slope power $k$ could expand available range of negative power of $\varepsilon$ for the fast learning convergence. Particularly, the negative power's threshold could be largely reduced within a narrow range of slope power $k \in [-0.1, 0)$. The terminated outputs would present somewhat different scale because of the unitary scalability. The above training is initialized with pure-imaginary number in the form of SVD decomposition of Gaussian random generator.

All the parameters are passive and fixed once the network is trained successfully, whereas the learned parameters are different from initialization. One of the tested results is presented in Figure 10(d), the illuminated coherent wavelength and diffraction distance retains unchanged in the forward tested stage. The diffractive distance and the coherent illumination wavelength could influence the random modulation for nodes or weight, presenting a great potential on recognition occasions with respect to distance. For exhibiting the function of phase-ReLu implemented in **[DN]$^2$,** Table 2 shows the learning performance with respect to different types of initialization in complex space. Separate type is good for phase-ReLu, and the imaginary part type is good for phase-LReLu.

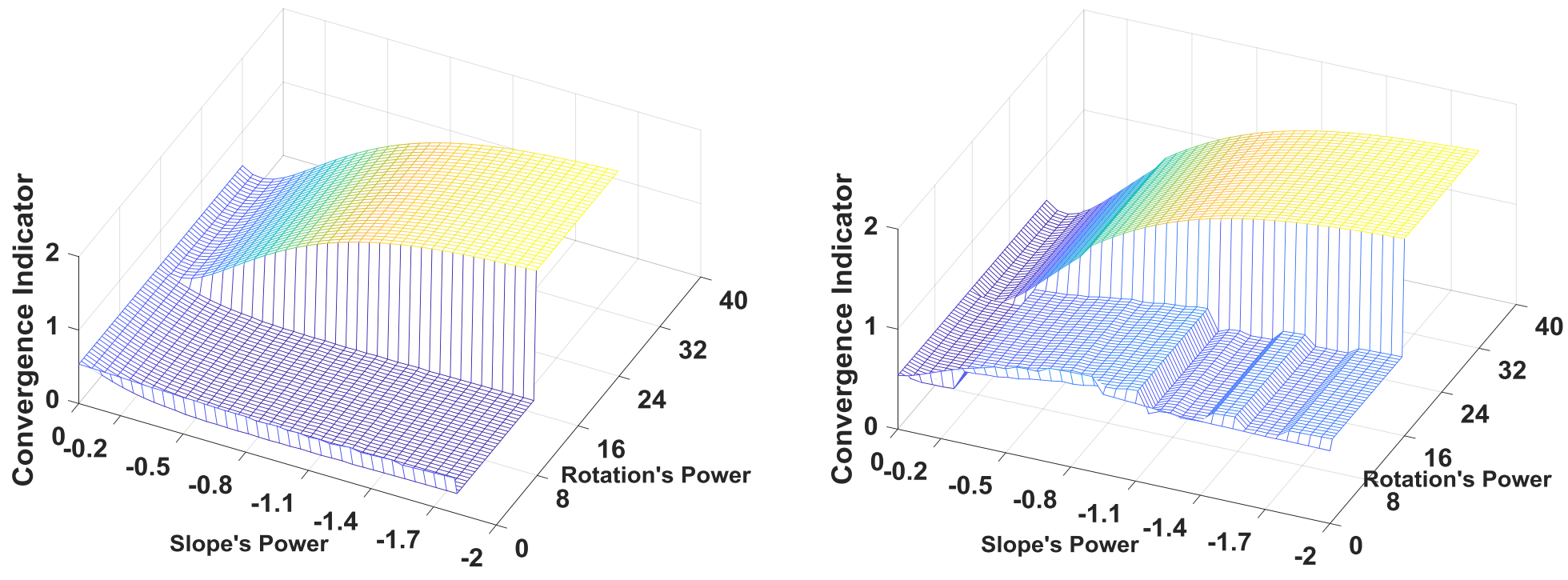

***Fig. 12*** The convergence enhancement by implementing rotation and slope tuning in pure-imaginary unitary initialization

(a) Clockwise rotation, (b) Anti-clockwise rotation

Table 2 (a) Learning performance in Unitary Learning with ReLu synapse

| | ReLu-Seperate | ReLu-Phase Only | ReLu-Imaginary |
|---|---|---|---|
| Rotation Plus | ✗ | ✗ | ✓ |
| Convergence | ✓ | ✓ | ✓ |
| Unitary Weight | ✓ | ✗ | ✗ |

Table 2 (b) Learning performance in Unitary Learning with LReLu synapse

| | LReLu-Seperate | LReLu-Phase Only | LReLu-Imaginary |
|---|---|---|---|
| Rotation Plus | ✗ | ✗ | ✓ |
| Convergence | ✓ | ✓ | ✓ |
| Unitary Weight | ✗ | ✗ | ✓ |

### 5.2 Deep diffractive neural network for 2D classification and verification

2D classification and verification are fundamental tasks operated in deep neural network and has achieved a fabulous accuracy with real-value backpropagation [34-35]. **[DN]$^2$** in complex space could provide high degree of freedom in wavelength, distance, phase for intelligent manipulation and modulation. We validate the unitary learning in **[DN]$^2$** for traditional binary and gray 2D classification and verification tasks-oriented, belonging to the situation that intensity-to-intensity nonlinear mapping with unitary initialization in complex space.

Firstly, a 2D classification on the MNIST dataset is investigated, sigmoid is chosen for judgment at output layer whereas tanh is chosen as the activation in hidden layer. The compatible condition enables sigmoid and tanh in complex space feasible in complex-valued Backpropagation. The feeding samples are 2D intensity images without any phase information. We also present three types of uniformly random initialization in complex space to observe the performance of unitary learning within two-layer **[DN]$^2$**. Some selected images from MNIST with a size of 28×28 pixels are displayed in Figure 13(a), and the number of entire samples involving training and validation sets is 60,000, and the learning rate is set as 0.2. The outputs practically present phase information in that the learned intelligent matrix and mask modulation cannot completely encode the input optical field into a pure-amplitude output optical field during the iterative epoch. Whereas, the trivial phase arising at the output layer could be ignored once the training is successful with backpropagation. As to test, it is rational for us to recognize the hand-written digital number through intensity because the incidental phase is very little. The initialization is implemented by Trabelsi formulation [12] that unitary matrix with a designed variance for amplitude and a uniform distribution for phase. The learning curves with Trabelsi initialization are displayed in Figure 13(b). The training are operated on the platform of Python version 3.6.7 using a desktop computer (Intel Core i7-4510U CPU at 2.60 GHz). The learning curves constantly descend during iterations, unitary learning is valid, even the performance is not very good as real space right now. The number of tested samples is 10,000, and the verification accuracy could achieve 92.64%. Figure 12(c) shows a pair of the training sample, the confusion matrix for the classification, and verification results whose values represent the accuracy rate. The labels could be indicated with the energy distributions focused on a specific region.

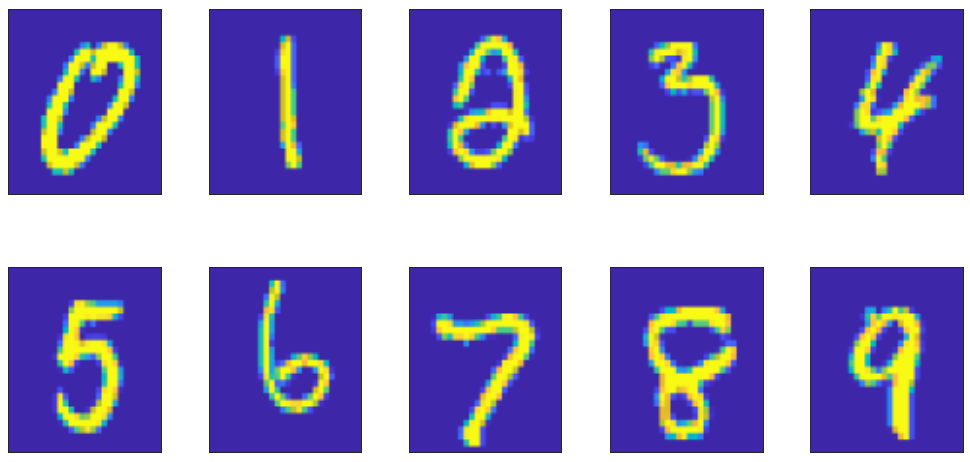

(a)

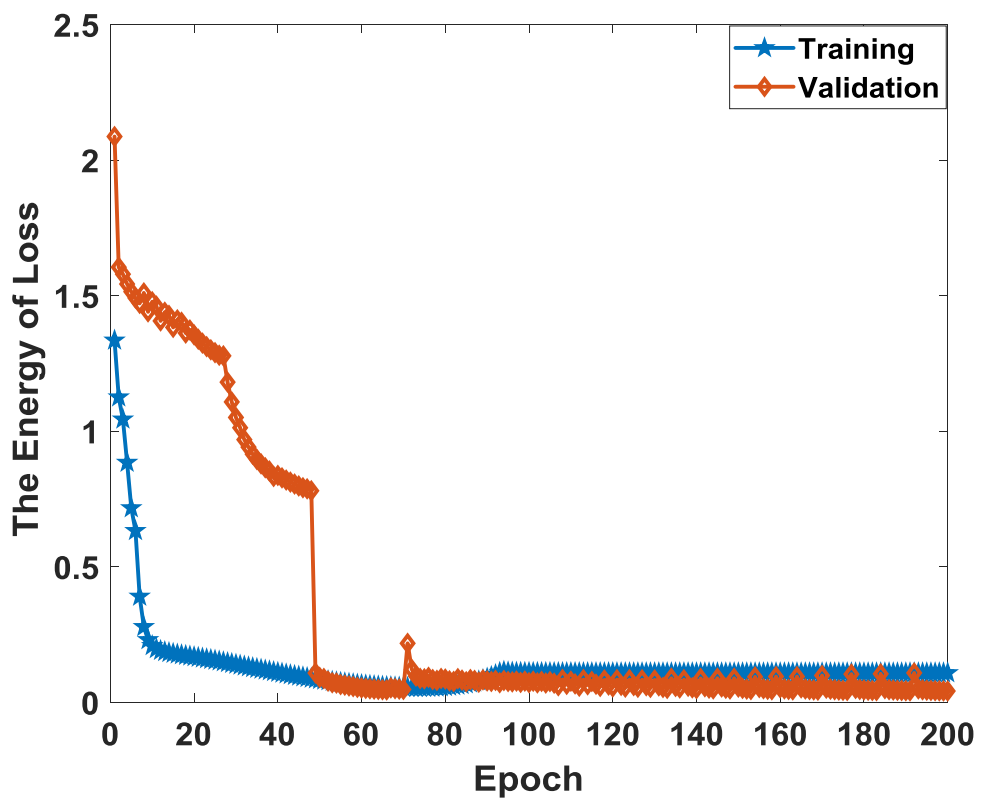

(b)

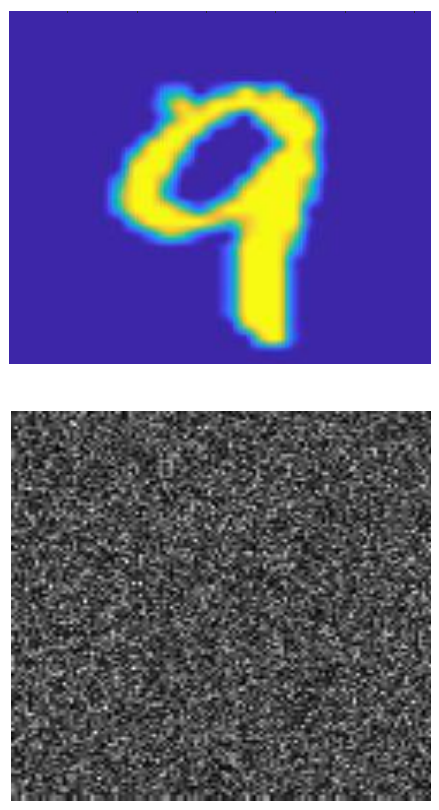

| 92.8 | 0.2 | 0.6 | 1.1 | 0.4 | 1.7 | 0.2 | 1.4 | 0.8 | 0.7 |
|---|---|---|---|---|---|---|---|---|---|
| 0.1 | 92.1 | 0.3 | 0.4 | 1.2 | 1.8 | 0.1 | 1.3 | 1.4 | 1.5 |
| 0.6 | 0.3 | 91.4 | 0.6 | 0.4 | 1.1 | 1.0 | 1.6 | 1.6 | 1.8 |
| 0.2 | 0.8 | 1.3 | 92.7 | 0.3 | 1.7 | 0.3 | 0.5 | 1.1 | 0.2 |
| 0.2 | 0.7 | 0.2 | 2.2 | 90.9 | 0.8 | 0.9 | 1.7 | 1.3 | 1.2 |
| 1.2 | 0.4 | 0.8 | 0.6 | 0.9 | 94.1 | 0.2 | 0.9 | 0.4 | 0.5 |
| 1.1 | 0.6 | 0.8 | 0.7 | 1.3 | 0.2 | 91.6 | 1.2 | 1.7 | 0.7 |
| 1.7 | 1.2 | 0.4 | 0.3 | 0.2 | 0.6 | 0.4 | 92.3 | 1.1 | 1.7 |
| 1.6 | 1.3 | 0.7 | 0.8 | 0.3 | 0.4 | 0.6 | 0.7 | 93.1 | 0.5 |
| 1.6 | 0.7 | 0.3 | 1.4 | 0.8 | 0.2 | 1.0 | 1.6 | 0.7 | 91.6 |

(c)

***Fig.13*** The classification and verification experiment on MNIST dataset with **[DN]$^2$**. **(a)** Selected samples exhibition; **(b)** Learning curves with tanh synapse and sigmoid output. (**c**) Confusion matrix for verification indicator.

Secondly, **[DN]$^2$** performed on the FASHION dataset continue to run, sigmoid is chosen for judgment at output layer whereas phase-EReLu is chosen as the synapse in hidden layer. The verification accuracy could achieve 85.87%. The aspects of FASHION are shown in Figure 14. The initialization and operational environment are the same as the situation in MNIST training and test, whereas, the verification accuracy has somewhat decreased compared to the binary classification and verification task on the MNIST dataset. A pair of training samples and their confusion matrix for the verified results are exhibited in Figure 14(c).

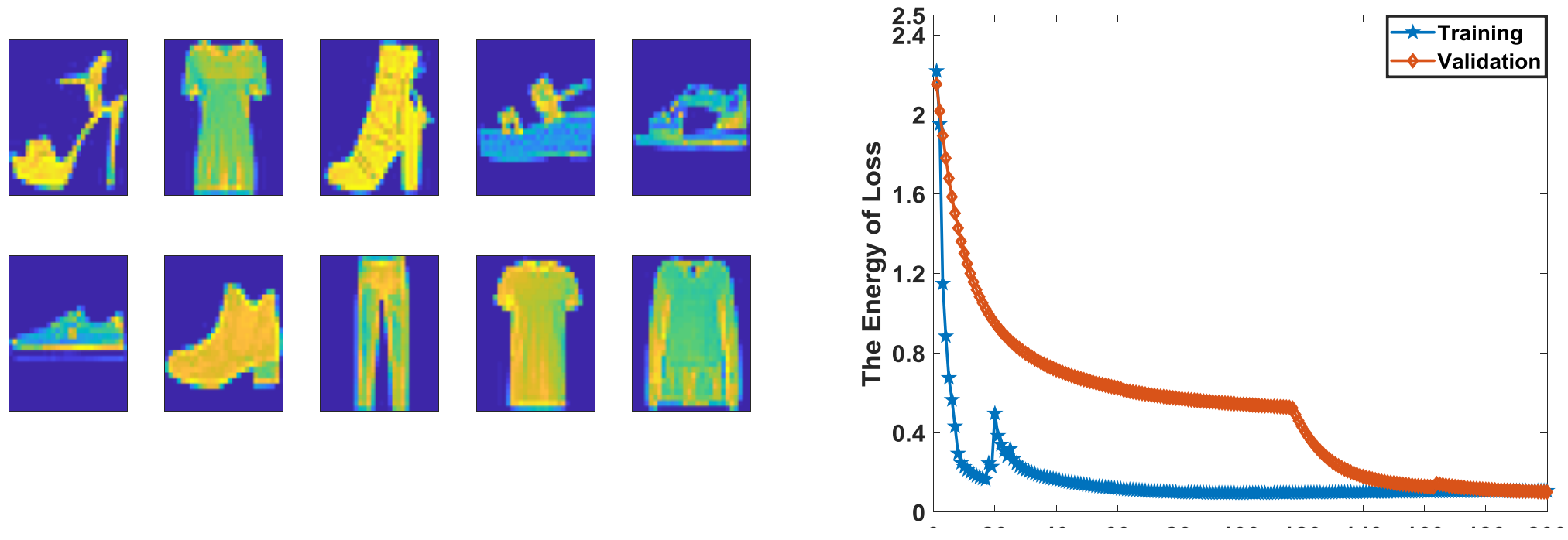

(a) (b)

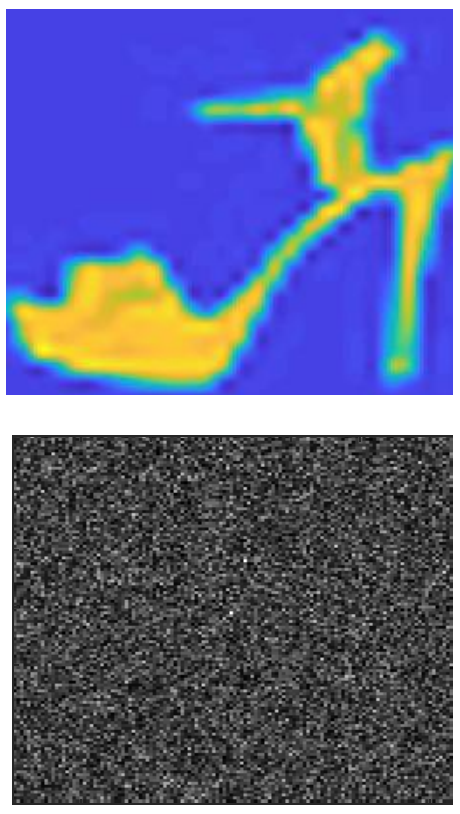

| 83.8 | 2.2 | 1.6 | 2.1 | 1.4 | 1.8 | 2.2 | 1.4 | 1.8 | 1.7 |
|---|---|---|---|---|---|---|---|---|---|
| 2.3 | 84.1 | 2.5 | 1.4 | 1.2 | 0.8 | 2.3 | 2.0 | 2.4 | 1.5 |
| 1.6 | 1.3 | 84.4 | 2.6 | 2.4 | 2.1 | 0.3 | 0.2 | 1.6 | 3.4 |
| 1.2 | 1.8 | 2.3 | 84.7 | 2.3 | 1.7 | 0.6 | 0.7 | 1.1 | 3.5 |
| 2.2 | 2.7 | 1.2 | 2.2 | 85.9 | 0.5 | 0.7 | 1.7 | 1.3 | 1.6 |
| 1.2 | 2.4 | 1.8 | 1.8 | 0.9 | 84.1 | 1.2 | 1.6 | 2.4 | 2.5 |
| 1.9 | 1.6 | 1.8 | 0.7 | 2.3 | 2.2 | 85.6 | 1.8 | 1.7 | 0.4 |
| 2.7 | 1.2 | 1.8 | 1.3 | 1.2 | 1.0 | 1.4 | 87.3 | 1.1 | 0.8 |
| 1.6 | 1.3 | 2.7 | 0.8 | 1.8 | 1.8 | 0.7 | 1.7 | 86.1 | 1.5 |
| 1.6 | 1.7 | 0.9 | 1.4 | 0.8 | 0.8 | 1.5 | 2.6 | 2.7 | 87 |

(c)

***Fig.14*** The classification and verification experiments on FASHION dataset with **[DN]$^2$**. **(a)** Selected samples exhibition; **(b)** Learning curves with phase-EReLu synapse and sigmoid output. (**c**) Confusion matrix for verification indicator.

Backpropagation prediction is different from phase retrieval for 2D classification and verification lying in its statistical ubiquity [36]. Firstly, the loss functions for optimization are built on different assumptions, the energy of complex optical field difference between forward output and target is implemented in backpropagation, and the loss function is represented as a multilayer implicit activation functions. Whereas, in phase retrieval, the energy error of optical intensity is implemented, and the multilayer structure in cascaded explicit format maps the input into targeted intensity without considering nonlinear activations. Secondly, the data-sources for iterations are provided in different modalities. Complex-value including amplitude and phase, regarded as a substantial entity, are fed to neural network in the format of batches, generalization or regression are learnable with backpropagation. Whereas, in phase retrieval, intensity constraint is enforced on a given pair of sample during iterations, discarding the phase equivalence at the output plane for the terminated convergence. The statistical inference and phase are lost in phase retrieval training.

## *8. Conclusions*

Deep diffractive neural network has a great superiority of parallel processing at speed of light, low power consumption as well as high bandwidth. In this article, the intrinsic unitary property in $[\mathbf{DN}]^2$ which is realized through diffractive modulation mechanism, is fully considered in training, called unitary learning. Unitary learning, covering its real counterpart, provides an alternative formulation for deep complex neural netowrk with unitary constraint on complex-valued weights. The adjoint compatible condition ensures that the fundamental sigmoid, tanh and quasi-Relu in complex space could be available as nonlinear activations with complex-variable conjugation substitution, whereas these pervasive activations have been empirically used, even rejected as activations in complex space for several decades. Unitary learning has a single channel competence in training, replacing the bumpy double channel training. It also provides alternative guidance for selecting suitable photoelectric materials as activations for photonic neural network Situ training considering both convergence and physical characteristics.

Unitary learning directly adopts the complex-value entity to update the weight, conjugated gradient descent is operated on Riemannian space, translating from Euclidean space with the help of an exponent map. Qausi-phase ReLu implemented in unitary learning is profoundly investigated for different application occasions, and some miraculous affections are revealed, such as phase break point for rotation.

Briefly, unitary learning reconciles the prolonged issues on activations selection in complex-valued backpropagation and unitary constraint on weight, stemming from the differentiability for nonlinear activations in the complex space. The new compact formulations could be easily extended with the concept of conjugation substitution, compared to the real-value backpropagation. To the best of our knowledge, it is launched for the first time. Two fundamental classification and verification tasks are operated with $[\mathbf{DN}]^2$, verifying the pragmaticality of unitary learning. Additionally, unitary learning renders a great potential in the photonic neural network meshed with M-Z nanophotonic components.

## *Disclosures*

The authors declare no conflicts of interest.

***Funding*** This work was supported by National Natural Science Foundation of China (61805208).